\documentclass[10pt,twocolumn,letterpaper]{article}
\usepackage[accsupp]{axessibility} 
\usepackage{cvpr}              

\usepackage{graphicx}
\usepackage{amsmath}
\usepackage{amssymb}
\usepackage{booktabs}
\usepackage{comment}
\newcommand{\ud}{\,\mathrm{d}}
\usepackage[pagebackref,breaklinks,colorlinks]{hyperref}
\usepackage[ruled]{algorithm}
\usepackage{algcompatible}
\usepackage[noend]{algpseudocode}
\usepackage[capitalize]{cleveref}
\crefname{section}{Sec.}{Secs.}
\Crefname{section}{Section}{Sections}
\Crefname{table}{Table}{Tables}
\crefname{table}{Tab.}{Tabs.}

\usepackage{enumitem}
\usepackage{amsmath}
\usepackage{graphicx}

\NewDocumentCommand{\codeword}{v}{%
\texttt{\textcolor{blue}{#1}}%
}

\def\cvprPaperID{7690} 
\def\confName{CVPR}
\def\confYear{2024}

\begin{document}

\title{Diffeomorphic Template Registration for Atmospheric Turbulence Mitigation}

\author{Dong Lao\\
UCLA Vision Lab \\
{\tt\small lao@cs.ucla.edu}
\and
Congli Wang\\
UC Berkeley\\
{\tt\small \!\!\!\!\!\! congli.wang@berkeley.edu}
\and
Alex Wong\\
Yale Vision Lab\\
{\tt\small alex.wong@yale.edu}
\and
Stefano Soatto\\
UCLA Vision Lab\\
{\tt\small soatto@cs.ucla.edu}
}

\maketitle

\begin{abstract}
We describe a method for recovering the irradiance underlying a collection of images corrupted by atmospheric turbulence. Since supervised data is often technically impossible to obtain, assumptions and biases have to be imposed to solve this inverse problem, and we choose to model them explicitly. Rather than initializing a latent irradiance (``template'') by heuristics to estimate deformation, we select one of the images as a reference, and model the deformation in this image by the aggregation of the optical flow from it to other images, exploiting a prior imposed by Central Limit Theorem. Then with a novel flow inversion module, the model registers each image TO the template but WITHOUT the template, avoiding artifacts related to poor template initialization. To illustrate the robustness of the method, we simply (i) select the first frame as the reference and (ii) use the simplest optical flow to estimate the warpings, yet the improvement in registration is decisive in the final reconstruction, as we achieve state-of-the-art performance despite its simplicity. The method establishes a strong baseline that can be further improved by integrating it seamlessly into more sophisticated pipelines, or with domain-specific methods if so desired. 
\end{abstract}

\section{Introduction}
\label{sec:intro}

\begin{figure*}[t!]
\centering
\includegraphics[width=0.9\textwidth]{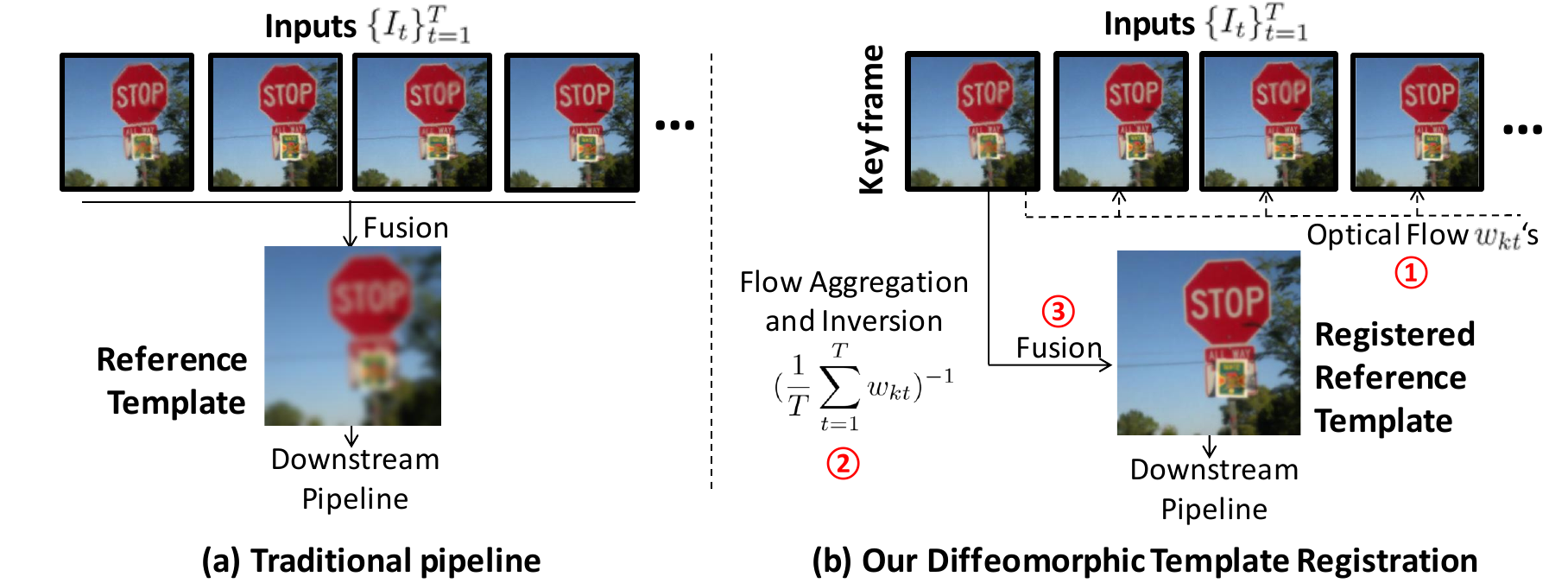}
\caption{\sl\small {\bf Schematic for Diffeomorphic Template Registration.} To mitigate atmospheric turbulence, one needs a ``reference template'' to estimate distortions. Existing methods use simple heuristics (e.g. averaging) to initialize this reference template and are susceptible to blurry artifacts due to a lack of registration. Instead, we select a keyframe and calculate warping from this keyframe to other frames by optical flow. By the Central Limit Theorem, the average of these optical flows converges to the warping from the keyframe to the ground truth reference template. With a novel flow inversion algorithm (detailed in Sect.~\ref{sec:inversion}), input frames can be registered to the reference template by Eq.~\eqref{eq:template}, even without explicit access to an initialized reference template. This registered reference template can seamlessly fit into existing downstream atmospheric turbulence mitigation pipelines, such as blind deconvolution and lucky image fusion.} 
    \label{fig:schematic}
\end{figure*}

Turbulent air motion, once temporally averaged and sampled to yield a digital picture, causes space- and time-varying blur. A model of the generative process for the measured data 
{\em given} the instantaneous ``true'' irradiance 
consists of simple integration over  the sampling time interval 
resulting in discrete samples. 
Inferring the true irradiance is an ill-posed problem, for there is an equivalence class of unknowns 
that yield the same measurements. 
We can therefore choose {\em any}  representative of the equivalence class, for instance one that would yield the sharpest solution. 
Traditionally the representative element is chosen as the (temporal) registered mean  
but since different images are blurred differently in different locations, the deformation 
is not identifiable at all locations, causing the representative 
to not have desirable properties, such as being blurry. Since the choice of representative is arbitrary, we can instead select the {\em sample} that has the most desirable properties, for instance the sharpest image (``lucky frame''), as a canonical reference from which to estimate the deformations. 
Thus, instead of solving the joint optimization for the deformations and the resulting average, 
we select the base frame $I_0 = I_\tau$ to correspond to a given image for some $\tau$, for instance one that  has the desired properties. Note that this choice is made without loss of generality for the purpose of estimating the deformations $\hat w_t$ mapping $I_t$ to $I_0$, since the choice of canonical representative in an equivalence class is arbitrary, so the mean has no more dignity than any of the samples. Once we have estimated the deformations, however, there is a choice to be made on what criterion to use to infer the approximation of the underlying ``true'' irradiance $I_0$. We stress that $I_0$ is not just unknown, but unknowable, and therefore disambiguation rests on the choice of prior, or model, or complexity criterion. 


Deep learning-based methods hide such arbitrary assumptions in the choice of architecture \cite{liang2021swinir,mehri2021mprnet,wang2022uformer,zamir2022restormer} (class of functions), or prior (loss function)  \cite{mao2022single,yasarla2021learning,yasarla2022cnn,jaiswal2023physics}. 
However, inductive prior based on data used for training can be problematic for discovery, where by definition one is seeking the unknown. Generic priors may therefore be better suited for applications to computational astronomical imaging.
Among non-learning methods, many directly estimate a reference image by using heuristics such as various forms of averaging \cite{shimizu2008super, zhu2012removing, lou2013video, mao2020image}, but mostly result in blurry reference images (Fig.~\ref{fig:template}). Computing deformation from this blurry reference image to other images severely impacts the quality of the deformation fields, which are sometimes of independent interest. We instead use an input image as the reference. By the Central Limit arguments, the mean of the deformation from this frame to other frames converges to the \emph{inverse} of the deformation from the ground truth to this frame (Eq.~\eqref{eq:inverse}). Although straightforward, this approach involves inverting the optical flow field, which can be prone to numerical instabilities. We tackle this flow inversion problem by hard-coding an interpolation scheme, a key contributor to the performance of our method (Sect.~\ref{sec:inversion}). 

The resulting method (Fig.~\ref{fig:schematic}) is simple and versatile since it is not subject to domain shift effects or inductive aberrations. To further emphasize the robustness of the method to the arbitrary choice of reference, in the experiments we simply use the first input frame as our reference frame. We also use a vanilla optical flow method, the good old-fashioned Horn \& Schunck ($L^2$-regularized) optical flow \cite{horn1981determining}. Yet, despite these simplistic choices, we achieve state-of-the-art results on atmospheric turbulence mitigation benchmarks. 
If one longs for more sophistication, our method can be seamlessly combined with any existing sharpest image selection, optical flow estimation, downstream deconvolution, lucky-image fusion, and dynamic object filtering scheme. Our main point is that we do not need any of the heuristics underlying these methods in order to obtain state-of-the-art results. 
Our processing pipeline is generic and can be improved by replacing any component with better ones, including domain-specific regularization if and when appropriate or desired.

\section{Related Works}
\label{sec:related}

Beyond hardware methods (\eg adaptive optics~\cite{wang2018megapixel}) to mitigate atmospheric turbulent distortion, there have been numerous software-based approaches. Some based on ``lucky image'' selection  \cite{anantrasirichai2013atmospheric,hardie2017block,law2009getting,mao2020image,lau2019restoration} choose a set of sharp images and compose them into a common frame; \cite{xie2016removing} employed deformation-guided spatial-temporal kernel regression to fuse the registered images together; \cite{he2016atmospheric} found the sharpest turbulence patches and enhanced them. To handle the blur artifacts within the composite frame, blind deconvolution~\cite{gilles2008atmospheric, he2016atmospheric,hirsch2010efficient,xie2016removing} is typically used, assuming the turbulence can be modeled as a spatially invariant linear filter \cite{lau2021atfacegan}. However, this assumption is often violated in atmospheric turbulence images due to the spatial-temporal changing displacements, leading to inaccurate results.

To address this issue, Zhu \etal~\cite{zhu2012removing} proposed registering each frame using B-spline based non-rigid registration to suppress geometric deformation, and then performs a temporal regression process to produce an image for blind deconvolution and generate the final output. 
\cite{gao2019atmospheric} used a convolutional neural network (CNN) to reconstruct turbulence-corrupted video sequence.
\cite{zhang2018removing} proposed a complex steerable pyramid framework to decompose the degraded image sequence.  \cite{anantrasirichai2023atmospheric} found that complex-valued convolutions better capture phase information than typical real-valued convolution.
Furthermore, \cite{li2021unsupervised} formulated geometric distortions removal as an unsupervised training step for a deep network, but must be repeated for each test image. \cite{wang2021deep} reconstructed the wavefront aberration phase from the distorted image of the object to perform  aberration correction for mitigating atmospheric turbulence. \cite{yasarla2021learning} used Monte Carlo dropout  toestimate uncertainty maps to guide image restoration; \cite{yasarla2022cnn} modeled the geometric distortion and blur at each pixel via variance maps and used to to aid the restoration of face images. In the domain of fluids, \cite{thapa2020dynamic} used recurrent layers to recover a mesh model of water surface from a monocular video, while \cite{li2018learning} undistorted dynamic refractive effects using warping network to remove geometric distortion and a color predictor net to further refine the restoration. Additionally, \cite{zamir2021multi,zamir2022restormer} performed a number restoration tasks, including denoising, deblurring, and dehazing. Although untested on atmospheric turbulence mitigation, the method appears promising. 

Recent studies~\cite{mao2020image,mao2022single} have proposed physics-inspired deep learning models that aim to capture the underlying physical processes that cause atmospheric turbulence. For example, TurbuGAN~\cite{feng2022turbugan} incorporates physically accurate simulation into the forward measurement model. These models leverage knowledge of the statistical properties of turbulence and use them to guide the restoration process.

However, it is difficult to obtain aligned clean and corrupted image pairs. Hence, there exists a very limited amount of publicly available datasets for imaging through atmospheric turbulence. Amongst them, the most widely used ones are limited to few tens of images \cite{anantrasirichai2013atmospheric,hirsch2010efficient,mao2020image,li2021unsupervised}. More recently, \cite{mao2021accelerating,mao2022single} proposed standardized benchmarks to evaluate the effectiveness of existing methods. Nonetheless, there remain challenges in collecting large-scale datasets for training. The majority of them, such as ~\cite{perlman2007data,li2008public}, simulate realistic atmospheric turbulence effects; yet, this process can be expensive. Another line of work follows optics and leverages split-step simulation, using ray-tracing and wave-propagation \cite{bos2012technique, hardie2017simulation,roggemann2018imaging,schmidt2010numerical}. The less expensive synthesis techniques rely
on random pixel displacement and a blur model \cite{chak2021subsampled,leonard2012simulation}, but are far from realistic.  To bridge this gap, \cite{chimitt2020simulating} simulated anisoplanatic turbulence by sampling intermodal and spatially correlated Zernike coefficients. \cite{chimitt2022real} further improved the speed of simulation to real-time by using a multi-aperture model \cite{chimitt2020simulating}, in conjunction with the phase-to-space transform \cite{mao2021accelerating}, and an approximation of the Zernike correlation tensor in order to provide a scalable platform for evaluation.

Overall, while there have been significant advances in software-based approaches for mitigating atmospheric turbulence, there are still many challenges that need to be addressed. These include accurately modeling the turbulence, developing efficient and effective deep learning-based methods, and creating large-scale benchmark datasets that accurately reflect real-world scenarios.

\section{Method}\label{sec:method}

We consider an image sequence $\{I_t\}_{t=1}^T$ with $I_t: \Omega \mapsto \mathbb{R}^n$, where $\Omega$ is the image domain and $n=1$ for gray-scale or $n=3$ for RGB color channels.

\def\figd{figures}
\def\fWidD{0.48\textwidth}
\begin{figure}[t]
\centering
\includegraphics[width=\fWidD]{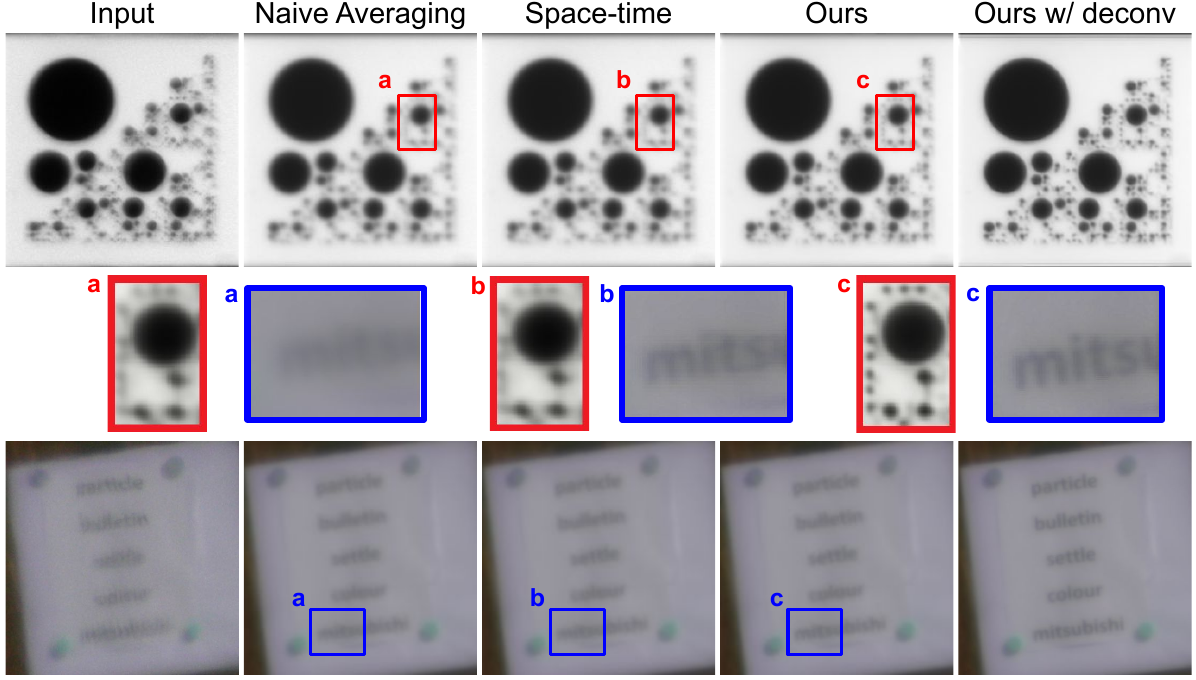}
\caption{\sl\small {\bf Sharper and more rigid reference template.} Compared with naively averaging and space-time non-local averaging \cite{mao2020image}, our method improves the quality of the reference template. Highlighted regions are zoomed-in and sharpened for the sake of visualization, better viewed in 3$\times$.}
\label{fig:template}
\end{figure}

\subsection{Modelling Deformation}

 We start from the standard Lambertian assumption, considering the case where there is no blur but only deformation exists in the sequence. Given the original ground truth image (without turbulence) $I$, for each time $t$, $I_t$ is deformed from $I$ by a diffeomorphic warping $w_t$, and 
\begin{equation}\label{eq:lambertian}
    I_t(w_t(x)) = I(x) + \eta_t(x), \quad x\in \Omega,
\end{equation}
where $\eta$ denotes i.i.d. zero mean Gaussian noise. We assume each $w_t$ is drawn from an i.i.d. distribution. With a stationary camera, by the Central Limit Theorem, $\frac{1}{T}\sum_{t=1}^T w_t \rightarrow \mathit{Id}$ as $T \rightarrow +\infty$ assuming the deformation is zero-mean, where $\mathit{Id}$ denotes the identity mapping. With Eq.~\eqref{eq:lambertian}, the turbulence mitigation problem can be written as 
\begin{equation}
    \underset{I, \{w_{t=1}^T\}}{\text{minimize}}\,\, \sum_{t=1}^{T} \int_{x \in \Omega} |I_t(w_t(x)) - I(x)|^2 \ud x,
\end{equation}
in which $I$ has an explicit optimizer 
\begin{equation}\label{eq:optimalI}
\hat{I}(x) = \frac{1}{T} \sum_{t=1}^T I_t(w_t(x)), \quad x\in \Omega ,
\end{equation}
and $w_t$ can be estimated by the optical flow from $I$ to $I_t$. 

However, without a good initialization of $I$, optical flow estimation for $w_t$ fails. Meanwhile, warping between two frames $w_{ij}$, on the other hand, can be estimated by the flow between $I_i, I_j$. By choosing a key frame $I_k$, through flow composition \cite{lao2017minimum,lao2018extending,lao2021flow}, we have $w_t$ = $w_{kt}\circ w_k$. In this way, we decompose the estimation of warping $w_t$ for \emph{each} $t$ into a composition of the warping $w_{kt}$ (directly estimated by optical flow) and a \emph{single} warping $I$ to each $I_k$.

\def\figd{figures/flow}
\def\fWidD{0.12\textwidth}
\begin{figure}[t]
\centering
{\footnotesize
\hspace*{-0.3in}
\begin{tabular}[0mm]{c@{\hskip 0.01in}c@{\hskip 0.01in}c@{\hskip 0.01in}c@{\hskip 0.01in}c}
&$I_0$&Flow $w_{01}$&$\alpha$ mask&Inverted: $w_{01}^{-1}$
\\ [1mm]
\rotatebox{90}{\,\,Flow Inversion}&\includegraphics[width=\fWidD]{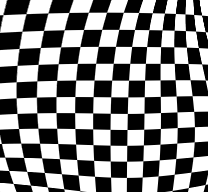}&
\includegraphics[width=\fWidD]{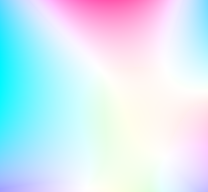}&
\includegraphics[width=\fWidD]{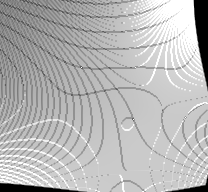}&
\includegraphics[width=\fWidD]{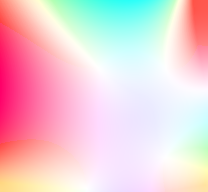}\\
\rotatebox{90}{\quad Registration}&\multicolumn{2}{c}{\quad\quad\quad\quad\includegraphics[width=0.13\textwidth]{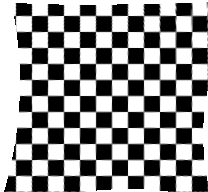}}&
\multicolumn{2}{c}{\hspace{-5em}\includegraphics[width=0.13\textwidth]{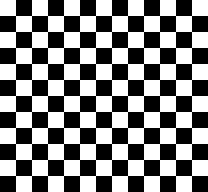}}\\
&\multicolumn{2}{c}{\quad\quad\quad\quad Warped $I_0(w_{01}^{-1}(\cdot)))$}&\multicolumn{2}{c}{\hspace{-5em}Ground Truth $I_1$}
\\
\end{tabular}
}
\caption{\sl\small {\bf Example of flow inversion and registration.} Given the flow $w_{01}$, we map the non-integer endpoints to the four nearest pixels, then construct an inverse flow $w_{01}^{-1}$ by weighted averaging. When the corresponding weight is zero in $\alpha$, we fill in the missing values by inpainting. This scheme successfully registers $I_0$ to the target frame $I_1$, even without having access to the ground truth $I_1$.}
\label{fig:flow}
\end{figure}

Recall $\frac{1}{T}\sum_{t=1}^T w_i \rightarrow \mathit{Id}$. By assuming a sufficiently large number of frames, for simplicity, we denote $\frac{1}{T}\sum_{t=1}^T w_i \approx \mathit{Id}$. Given the same reference frame $I_k$, one can combine all $w_{kt}$'s by 
\begin{equation}\label{eq:clt}
\mathit{Id} \approx \frac{1}{T}\sum_{t=1}^T w_t = \frac{1}{T}\sum_{t=1}^T w_{kt}\circ w_k =(\frac{1}{T}\sum_{t=1}^T w_{kt})\circ w_k .
\end{equation}
Therefore, 
\begin{equation}\label{eq:inverse}
 w_k \approx (\frac{1}{T}\sum_{t=1}^T w_{kt})^{-1},
\end{equation}
where the exponent denotes the inverse of the diffeomorphic warping. Note that, $w_k$ is estimated with a combination of optical flows across input frames only, \emph{without} requiring access to an explicitly initialized template. Computing Eq.~\eqref{eq:inverse} is non-trivial, as inverting a diffeomorphic warping (parametrized by 2-D optical flow) has no explicit solution. We describe how we implement this function in the Sect.~\ref{sec:inversion}. Finally, by Eq.~\eqref{eq:optimalI} and Eq.~\eqref{eq:inverse}, we have 
\begin{equation}\label{eq:template}
\hat{I}(x) \approx \frac{1}{T} \sum_{t=1}^T I_t(w_{kt}\circ (\frac{1}{T}\sum_{t=1}^T w_{kt})^{-1}(x)), \quad x\in \Omega ,
\end{equation}
where we name $\hat{I}(x)$ the \emph{template} of the sequence.

\subsubsection{Flow Inversion}\label{sec:inversion}

Inverting optical flow seems natural but is seldom practiced. When dealing with simple flow patterns like affine motion, an explicit solution is easy to compute in closed form. However, for more complex scenarios, such solutions are generally ill-posed. When pixels from one frame are mapped to non-integer positions in another frame, simply reversing the direction in the target frame is not a practical approach. Moreover, issues like occlusions at image boundaries and motion boundaries can result in missing values at certain pixel positions. One notable approach to inverting optical flow is offered by SMURF \cite{stone2021smurf}, which proposes learning of a flow inverting network on the fly. However, this method is not suitable for atmospheric turbulence removal problems with limited computational resources and processing time.

In this study, we propose a solution for estimating flow inversion, which is detailed in Alg.~\ref{alg:inversion}. Our approach involves determining the endpoint for each pixel in the reference frame and mapping the inverse flow from non-integer endpoints in the target frame to the four nearest integer points where pixels are situated. Since multiple flow vectors may map to the same target pixel, we calculate a weighted average to derive the inverse flow. For pixels with zero weights, we fill in the missing values in the inverse flow using spatial interpolation (inpainting). An example is in Fig.~\ref{fig:flow}. Even without having access to the true target $I_1$, given only the forward flow $w_{01}$, this scheme successfully inverts the flow and registers $I_0$ to the target frame.

The computation time for a $256\times256$ frame on a laptop CPU in MATLAB is less than 0.02 seconds, and with parallelization, we anticipate significant speed enhancements. This suggests that our approach has the potential to be applied in real-time computational imaging applications.

\renewcommand{\algorithmiccomment}[1]{\hfill$\triangleright$\textit{#1}}
\begin{algorithm}[t]
\small
  \begin{algorithmic}[1]
    \Require Input flow field $w \in \mathbb{R}^{h\times w\times 2}$
    \State Initialize: $w^{-1} = \text{zeros}(h,w,2)$, $\alpha = \text{zeros}(h,w)$
    \For {each grid point $(i,j)$ in the image domain, } 
    \State Compute endpoint $(\hat{i},\hat{j}) = (i,j) + w(i,j)$
    \State Find four neighboring grid points of $(\hat{i},\hat{j})$: $\mathcal{N}(\hat{i},\hat{j}) = \{\text{floor}(\hat{i},\hat{j}),\text{floor}(\hat{i}+1,\hat{j}),\text{floor}(\hat{i},\hat{j}+1),\text{floor}(\hat{i}+1,\hat{j}+1)\}$
    \For {each $(i', j') \in \mathcal{N}(\hat{i},\hat{j})$} 
         \State $w^{-1}(i',j')\mathrel{+}=- w(i,j)\times(2-|(\hat{i},\hat{j}) - (i',j')|^1_1)$
            \State  $\alpha (i',j')\mathrel{+}=(2-|(\hat{i},\hat{j}) - (i',j')|^1_1)$
    \EndFor
        \EndFor
            \For {each grid point (i,j) in the image domain, }
            \If  {$\alpha(i,j)\neq 0$}
            $w^{-1}(i,j)=w^{-1}(i,j)/\alpha(i,j)$ \EndIf
            \EndFor
    \State Inpaint $w^{-1}(i, j)$ by interpolation where $\alpha(i,j)=0$\\
    \Return Inverse flow $w^{-1}$
    
  \end{algorithmic}
  \caption{\sl Optical Flow Inversion.}
  \label{alg:inversion}
\end{algorithm}

\subsection{Modeling Blur}
Temporal atmospheric turbulence causes wavefront distortion, resulting in image blurring. The image formation model has been discussed extensively in~\cite{chan2022tilt}. Following derivation shows that the per-frame degradation point spread function (PSF) can be approximated as a parametric Gaussian function. 

\subsubsection{Point Spread Function}
In Fourier optics~\cite{goodman2005introduction}, the time-varying PSF $p_t(x)$ of an ideal imaging system with an effective focal length~$f$ at wavelength $\lambda$ can be expressed as:
\begin{equation}\label{equ:psf}
    p_t(x) = |h_t(x)|^2, \quad 
    h_t(x) = \mathcal{F} \left\{ A(u) e^{i W_t(u)} \right\}\left( \frac{x}{\lambda f} \right).
\end{equation}
where $A(u)$ represents the camera's entrance pupil at aperture coordinate $u$, $W_t(u)$ is the instantaneous wavefront that varies with time, and $\mathcal{F}$ denotes the Fourier transform. Assuming non-dispersivity, the wavefront remains unchanged with respect to the spectrum wavelength $\lambda$, and we will drop $\lambda$ in the followings.

\subsubsection{Turbulence Wavefront Decomposition}

\def\fWidD{0.24\columnwidth}
\begin{figure}
\centering
\footnotesize
\setlength{\tabcolsep}{0.3mm}
\begin{tabular}{cccc}
\includegraphics[width=\fWidD]{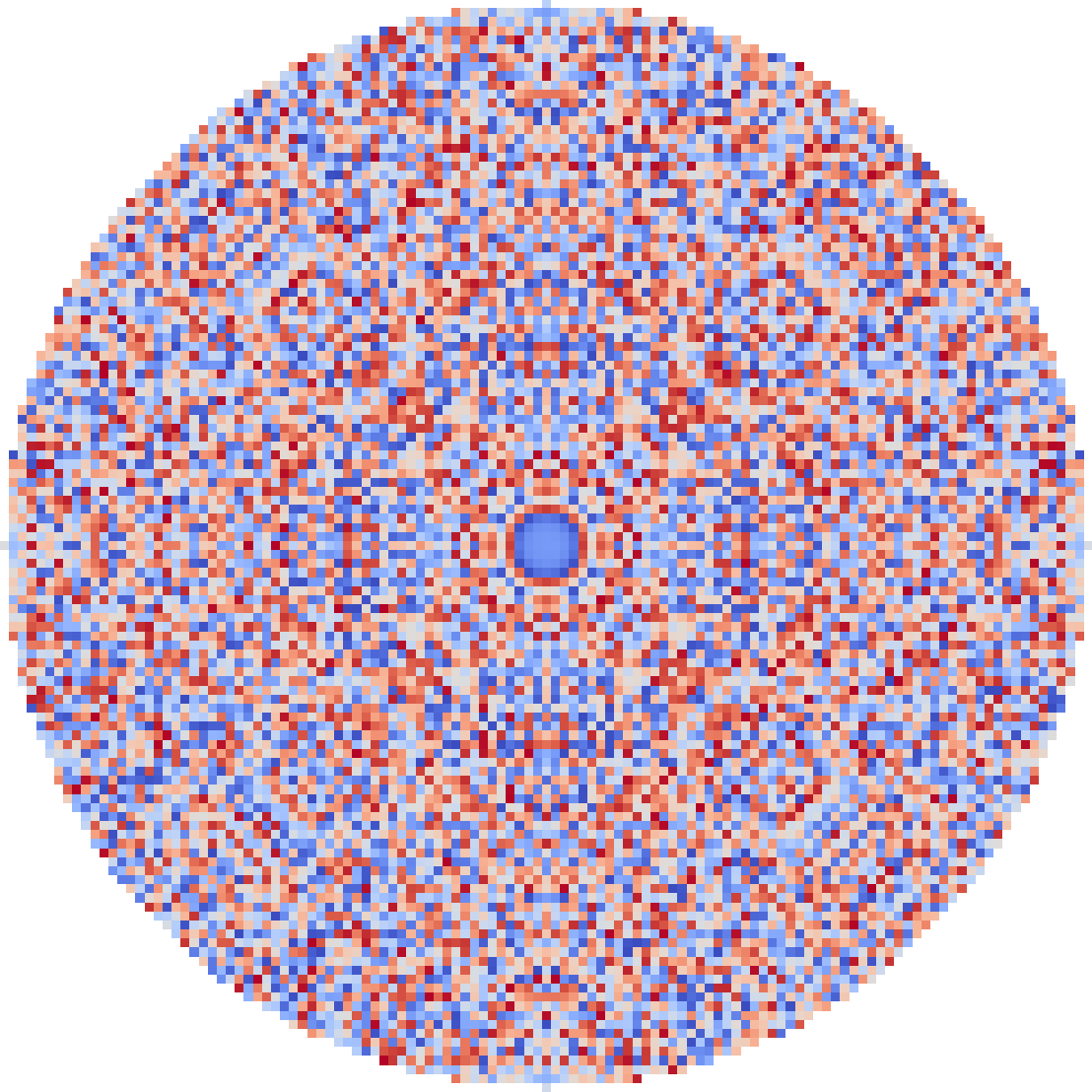} &
\includegraphics[width=\fWidD]{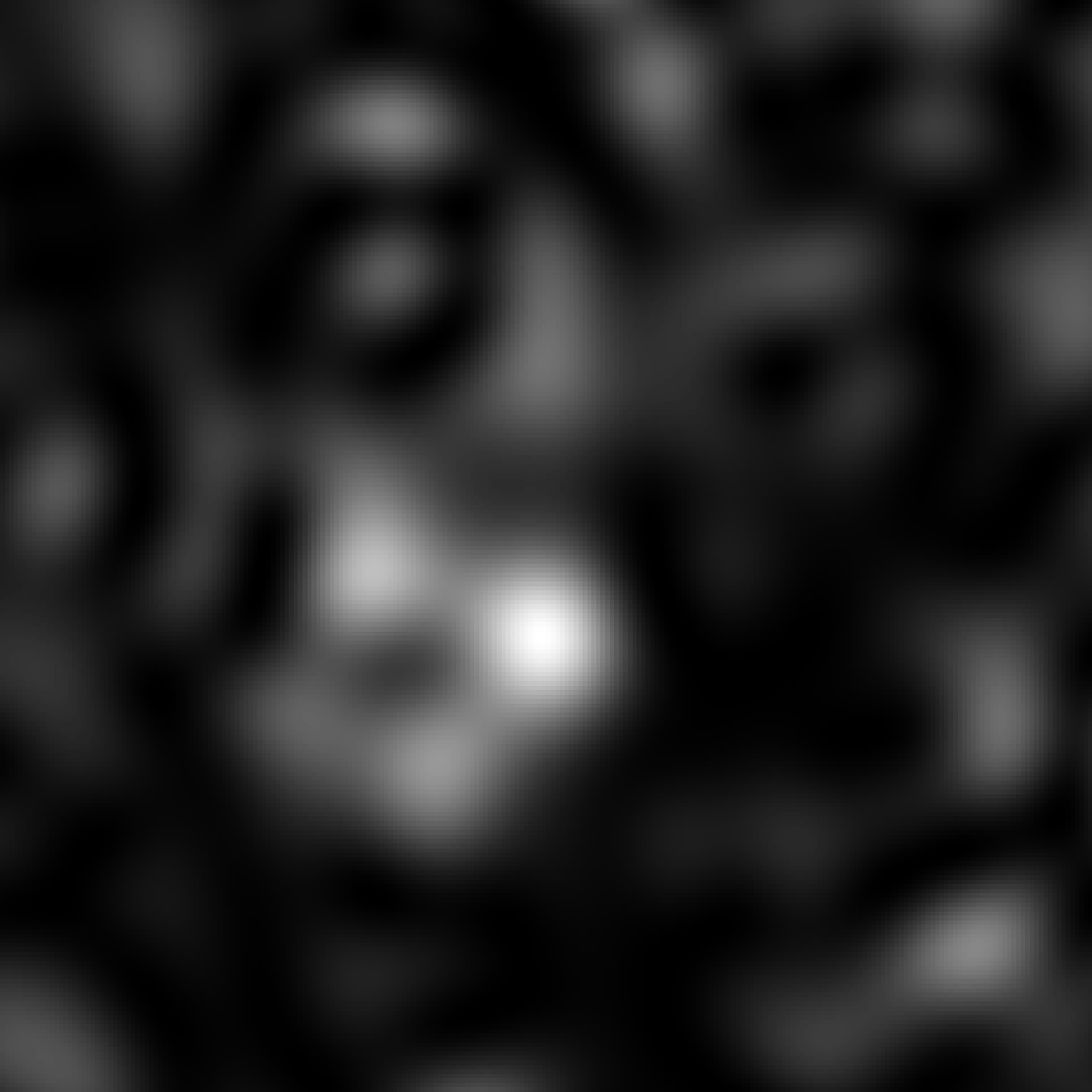} &
\includegraphics[width=\fWidD]{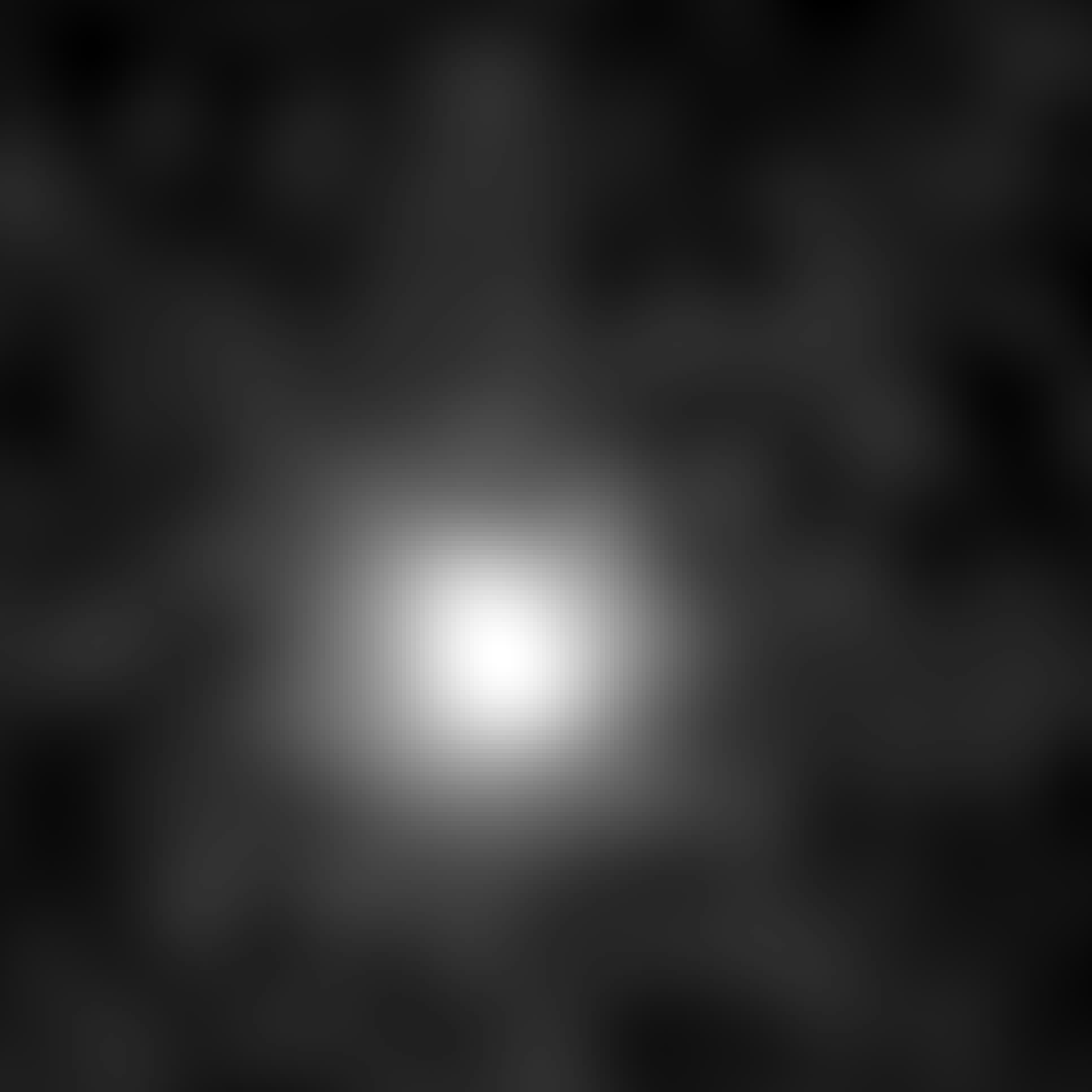} &
\includegraphics[width=\fWidD]{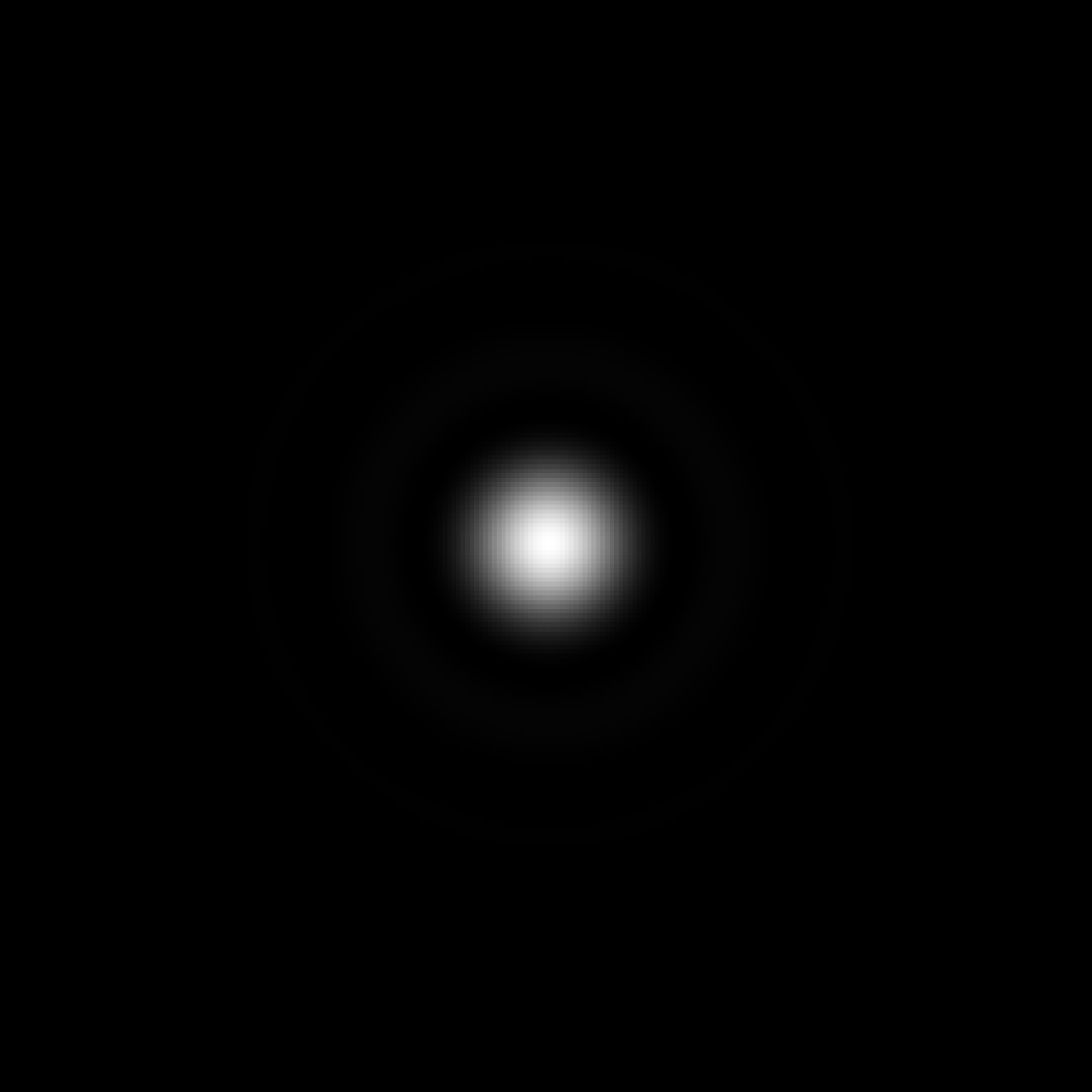} \\ [0.5mm]
Instant. $W_t(u)$ & Instant. $p_t(x)$ & Per-frame $\bar{p}(x)$ & $J_{\text{airy}}(x)$
\end{tabular}
\caption{\sl\small\textbf{Approximating blurring kernels as Gaussians.} Given a set of instantaneous wavefronts (visualized within a modulo of $2\pi$) and their PSFs, the per-frame PSF is approximated as a Gaussian function, characterized by shifts and a standard deviation.}
\label{fig:airy_disc}
\end{figure}

Wavefront $W_t(u)$ is typically decomposed into polynomial orders, \eg the Zernike polynomials~\cite{noll1976zernike}. Here we are most interested in the following decomposition:
\begin{equation}\label{equ:wavefront}
W_t(u) \quad = 
\underbrace{a_t u}_{\text{1st order}} + 
\underbrace{b_t u^2}_{\text{2nd order}} + 
\underbrace{c_t  V_t(u)}_{\geq \text{3rd orders}},
\end{equation}
where $a_t$, $b_t$, and $c_t$ are time-varying parameters due to turbulence, and $V_t(u)$ contains higher-order residuals.

\subsubsection{Blurring Kernels Approximated as Gaussians}
By substituting Eq.~\eqref{equ:wavefront} into the kernel $h_t(x)$ in Eq.~\eqref{equ:psf} and utilizing the convolution theorem with $\otimes$ being the convolution operator, we can rewrite $h_t(x)$ as follows:
\begin{align}\label{equ:h_decomposition}
    h_t(x)
    & =
    \mathcal{F} \left\{ A(u) e^{i a_t u}e^{i b_t u^2}e^{i c_t V_t(u)} \right\}
    \nonumber\\
    & =
    \mathcal{F} \left\{ e^{i a_t u} \right\} \otimes
    \mathcal{F} \left\{ A(u) e^{i b_t u^2} \right\} \otimes
    \mathcal{F} \left\{ e^{i c_t V_t(u)} \right\}
    \nonumber\\
    & = 
    g_{\text{airy}}(x - \Delta x_t; \sigma_t) \otimes
    \mathcal{F} \left\{ e^{i c_t V_t(u)} \right\},
\end{align}
where it is assumed that $A(u)$ is a circular aperture, as in most situations. Optically, $g_{\text{airy}}(x - \Delta x_t; \sigma_t)$ is the spatially-shifted ($\Delta x_t$), defocused ($\sigma_t$), amplitude Airy disc diffraction pattern of the imaging system, and it is a real number due to symmetry regarding $u$. By assuming in Eq.~\eqref{equ:wavefront}:
$$
c_t \sim \mathcal{N}(0, \sigma_c^2)
\quad\text{and}\quad
V_t(u) \ll 1,
$$
we obtain:
\begin{equation}
    \mathcal{F} \left\{ e^{i c_t V_t(u)} \right\} \approx 
    \delta(x) + i c_t \mathcal{F} \left\{ V_t \right\}(x).
\end{equation}
This allows rewriting Eq.~\eqref{equ:h_decomposition} to approximate $p_t(x)$ as:
\begin{align}\label{equ:p_approximation}
    p_t(x)
    & = |h_t(x)|^2 = \Re\{ h_t(x) \}^2 + \Im\{ h_t(x) \}^2
    \nonumber\\
    & \approx J_{\text{airy}}(x - \Delta x_t; \sigma_t) + o_t(x)
    \nonumber\\
    & \approx J_{\text{airy}}(x - \Delta x_t; \sigma_t),
\end{align}
because $o_t(x)$ is small, where:
$$
\begin{aligned}
    J_{\text{airy}}(x - \Delta x_t; \sigma_t)
    & = \left| g_{\text{airy}}(x - \Delta x_t; \sigma_t) \right|^2
    \\
    o_t(x)
    & = |c_t|^2 \cdot \left| g_{\text{airy}}(x - \Delta x_t) \otimes
    \mathcal{F} \left\{ V_t \right\}(x) \right|^2.
\end{aligned}
$$
$J_{\text{airy}}(x - \Delta x_t; \sigma_t)$ is the intensity Airy disc function, with defocus $\sigma_t$ and shift $\Delta x_t$. For each frame, the per-frame PSF is the aggregation of multiple instantaneous PSF $p_t(x)$, and thus we can approximate it as a Gaussian function, of mean $\mu$ and standard deviation $\sigma$:
\begin{equation}\label{equ:psf_gaussian}
\bar{p}(x) = \sum_{t \in \text{frame}} p_t(x) \approx G(x - \mu; \sigma).
\end{equation}
The approximation in Eq.~\eqref{equ:psf_gaussian} is illustrated in Fig.~\ref{fig:airy_disc}.


\subsection{Blind Deconvolution of A Uniform Kernel}
The degradation of each frame can thus be described as:
\begin{equation}\label{equ:main_deblur}
    \hat{I}(x) \approx \bar{p}(x) \otimes I(x).
\end{equation}
With the proposed diffeomorphic template registration, the template of the sequence is obtained, and Eq.~\eqref{equ:main_deblur} is solved as a blind deconvolution problem. The problem is typically addressed using total variation (TV) regularization or other possible regularizers. The objective is to minimize:
\begin{equation}
    \underset{I}{\text{minimize}}\,\, \int_{x\in \Omega} |\hat{I}(x) - \bar{p}(x) \otimes I(x)|^2 + \alpha |\nabla I(x)| \ud x,
\end{equation}
where $\alpha > 0$ is a tradeoff parameter. The problem is trivially solved using automatic differentiation engines.

There are two sources of blur in the template. The first source is caused by optical flow end-point error, which is typically assumed to be a zero-mean Gaussian distribution. With a sufficiently large number of frames, this results in a Gaussian blur being applied to the image. It can be proven that the two sources share the same Gaussian kernel.
The second source of blur is the inherent blurring caused by atmospheric turbulence, which cannot be explicitly measured. Other methods simply apply a training loss. By the Central Limit Theorem, with a sufficiently large number of frames, we assume that the blur adds up to a Gaussian distribution.


\subsection{Implementation Details}\label{sec:implementation}
\begin{algorithm}[t]
\small
  \begin{algorithmic}[1]
    \Require Input image sequence $\{I_i, i = 1,2, \dots, t\}$
    \State Choose the reference frame $I_{\rm ref}$
    \For {$i = 1, \dots, t$} \Comment{Can be parallelized} \label{step1}
    \State Estimate $w_i = \text{Flow}(I_{\rm ref}, I_i)$ \EndFor
    \State Compute $\bar{w} = \frac{1}{t}\sum_{i=1}^{t} w_i$
    \State Compute $\bar{w}^{-1}$ by Alg.~\ref{alg:inversion}
    \For {$i = 1, \dots, t$}
    \State Compose flow: $\hat{w_i} = \bar{w}^{-1} \circ w_i$\ \EndFor
    \State Compute $I_{\rm template} = \frac{1}{t}\sum_{i=1}^{t} \hat{w_i}^{-1}(I_i)$ 
    \State Blind deconvolution $I = \text{Deconv} (I_{\rm template})$ \label{step2}
    \State (Optional) $I_{\rm ref} = I$, repeat Step \ref{step1} to \ref{step2} \label{step3}
    \State Return $I$
    
  \end{algorithmic}
  \caption{\sl Full Algorithm.}
  \label{alg:full}
\end{algorithm}

The full algorithm pipeline is presented in Alg.~\ref{alg:full}. Our algorithm offers flexibility in choosing the optical flow for estimating warpings. In our experiments, we use Horn-Schunk \cite{horn1981determining} optical flow as a default choice, unless specified otherwise. The computation of optical flow is efficiently parallelized using MATLAB's \codeword{parfor} function. Flow composition and image interpolation follow~\cite{lao2018extending}.

One notable advantage of our approach is that it does not require a predefined reference template, so one is free to select any frame from the image sequence as the keyframe. In our experiments, we simply use the first frame for demonstration. However, one can also choose the sharpest image in the sequence as the reference frame or iterate by using intermediate results as in Step \ref{step3}.

For blind deconvolution, we have incorporated the ADMM blind deconvolution method from the public codebase in \cite{mao2020image}. The template then be integrated with a majority of turbulence mitigation pipelines. We demonstrate this by combining our method with the lucky image fusion module of \cite{mao2020image}, as in Sect.~\ref{sec:results}. A demo code is available in the supplementary material. 

\section{Experiments}\label{sec:results}
\textbf{Experimental Settings.}
Due to the nature of the task of atmospheric turbulence mitigation, there is a limited choice of datasets with available ground truth for quantitative evaluation. We employed two datasets: HeatChamber \cite{mao2022single} and CLEAR-sim \cite{anantrasirichai2013atmospheric}. HeatChamber contains 200 sequences of 90 images each at $220 \times 220$ pixel resolution, along with corresponding ground truth images. It was captured by displaying ground truth images on a screen 20 meters from the camera while inducing controlled turbulence using heat sources. To our knowledge, HeatChamber is the only real-world public dataset that provides ground truth for this purpose. In contrast, CLEAR-sim is a simulation dataset with 23 image sequences, featuring resolutions ranging from $256 \times 256$ to $1024 \times 1024$ pixels. Each sequence contains 96 to 100 frames with ground truth images.

We evaluate the results on both datasets using standard reconstruction metrics: Peak Signal-to-Noise Ratio (PSNR) and Structural Similarity Index (SSIM). Additionally, we present quantitative results on datasets, such as the one introduced by~\cite{hirsch2010efficient}, where ground truth is not available.

In our experimental comparisons, we focus on two representative methods: model-based methods TurbRecon \cite{mao2020image}, and learning-based method NDIR~\cite{li2021unsupervised}. TurbRecon generates a reference template through space-time non-local averaging, followed by registration using optical flow and lucky image fusion. For a fair comparison, we adopt the same optical flow method \cite{liu2009beyond} as TurbRecon, and kept all the hyper-parameters consistent with it. NDIR models non-rigid distortions as deformable grids and utilizes deep neural networks to optimize both the grids and the underlying clean image simultaneously. NDIR initializes the reference template using the mean of the input images.

\begin{table}[t]
\setlength{\tabcolsep}{1.4pt} 
  \centering
  \begin{tabular}{l|c|c|l|c|c}
  \multicolumn{3}{c|}{Single Frame}&\multicolumn{3}{c}{Multi Frame}\\
       \hline
       Method&PSNR&SSIM&Method&PSNR&SSIM\\    
    \hline
    TRDN\cite{yasarla2021learning}&18.42&0.642&CLEAR\cite{anantrasirichai2013atmospheric}&18.91&0.757\\    
    Uformer\cite{wang2022uformer}&18.68&0.658&NDIR\cite{li2021unsupervised}&20.08&0.763\\
Restormer\cite{zamir2022restormer}&19.12&0.684&Ours-template&20.30&0.780\\    
TurbNet\cite{mao2022single}&19.76&0.693&TurbRecon\cite{mao2020image}&20.62&0.787\\
PiRN\cite{jaiswal2023physics}&20.59&0.712&Ours-lucky&\bf{20.74}&\bf{0.789}\\

    \end{tabular}
  \caption{\sl\small{\bf Comparison on HeatChamber dataset.} While single-image methods can attain competitive PSNR, they exhibit low SSIM due to limited ability to handle deformation. Among multi-frame methods, our reference template (Ours-template) surpasses NDIR even without additional downstream processing. When integrated with the lucky image fusion pipeline (Ours-lucky), our approach enhances TurbRecon in both PSNR and SSIM metrics.}
     \label{tab:heatchamber}
\end{table}

\textbf{Quantitative Results.}
We present results on HeatChamber in Tab.~\ref{tab:heatchamber} and also compare with five deep-learning-based single-image reconstruction methods. Although single-image approaches can achieve competitive PSNR due to their capability to generate realistic reconstructions, their effectiveness is limited in capturing deformation. This limitation results in generally poor SSIM scores, as inferring deformation from a single image is suboptimal.

\begin{figure}
\vspace{-2mm}
\begin{center}
\vspace*{-10mm}
\begin{minipage}{.5\textwidth}
\vspace*{10mm}
\hspace*{-3mm}
   \includegraphics[width=0.47\columnwidth]{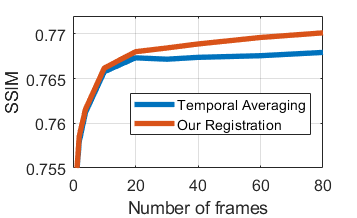}
   \caption{\sl\small\textbf{Sensitivity to the number of input frames.} Left: Our method yields improvements over temporal averaging  with respect to the number of frames. Right: Using 10 frames as input, ours outperforms NDIR in PSNR and SSIM.}
   \label{fig:multi-frame}
\end{minipage}
\begin{minipage}{0.5\textwidth}
\vspace*{-10mm}
\hspace*{-5cm}
\begin{tabular}{c|cc}
    \hline
    & NDIR & Ours \\
    \hline
    PSNR & 20.08 & \textbf{20.28} \\
    SSIM & 0.763 & \textbf{0.770} \\
    \hline
    Time (sec.) & 1800 & \textbf{16} \\
    \hline
\end{tabular}

\end{minipage}
\end{center}
\vspace{-4mm}

\end{figure}

In comparison to multi-frame methods, our reference template (referred to as Ours-template) outperforms NDIR even without any subsequent processing steps. Notably, the SSIM exceeds that of NDIR by 0.018, indicating that the diffeomorphic registration scheme better captures the scene's structure. We also test our method with a range of input frames as a sensitivity study (Fig.~\ref{fig:multi-frame}), where we consistently improve over temporal averaging and also NDIR for 10 input frames.  As stated in Sect.~\ref{sec:implementation}, our template seamlessly integrates into a majority of existing pipelines. By replacing TurbRecon's space-time non-local averaging scheme with our approach and adopting its lucky image fusion pipeline, the results (labeled as Ours-lucky) also demonstrate improvements over the original TurbRecon in both PSNR and SSIM metrics. In addition, we present results comparing our approach with NDL~\cite{zhu2012removing}, which also suggests a registration-deconvolution process. However, they employ a different registration strategy by using the average of input frames as a reference template and utilizing B-spline for deformation modeling instead of optical flow. Due to the extensive computational time (exceeding 1 hour for a $220\times220$ gray-scale image) required by NDL, we conduct comparisons on gray-scale images. Our method enhances the PSNR from 20.65 to 20.84 and SSIM from 0.715 to 0.723, while achieving a speed improvement of $\sim 100\times$.

\begin{table}
\begin{center}
    \begin{tabular}{c|ccc}
    \hline
    Key frame & Unlucky & Arbitrary ($1^{\text{st}}$ frame) & Lucky
    \\
    \hline
    SSIM & 0.770 & 0.771 & 0.771
    \\
    \hline
    \end{tabular}
    \label{tab:ref_frame}
\end{center}
    \vspace{-5mm}
    \caption{\sl\small\textbf{Sensitivity to frame selection.} We evaluate the registered template (before deconvolution) using the sharpest (``lucky'') and blurriest (``unlucky'') and show that our method is not sensitive.}
    \vspace{-0mm}
\end{table}

Our experiments on the CLEAR-sim dataset consistently support our prior findings. It is important to note that CLEAR-sim and HeatChamber differ in resolution. Utilizing an Nvidia GTX 1080Ti GPU with 11 GB of memory, our experiments reveal that NDR, can only process 10 input frames at a low spatial resolution of $220\times220$. In contrast, both our approach and TurbRecon, which employ optical flow to explicitly address deformation, can handle the full resolution of $1024\times1024$. This highlights the versatility of leveraging optical flow, eliminating the need for joint optimization that is computationally expensive. The quantitative results align with those from HeatChamber, demonstrating that TurbRecon outperforms NDR, with our method yielding the best performance among the three approaches.

\begin{figure*}[t!]
\centering
\includegraphics[width=\textwidth]{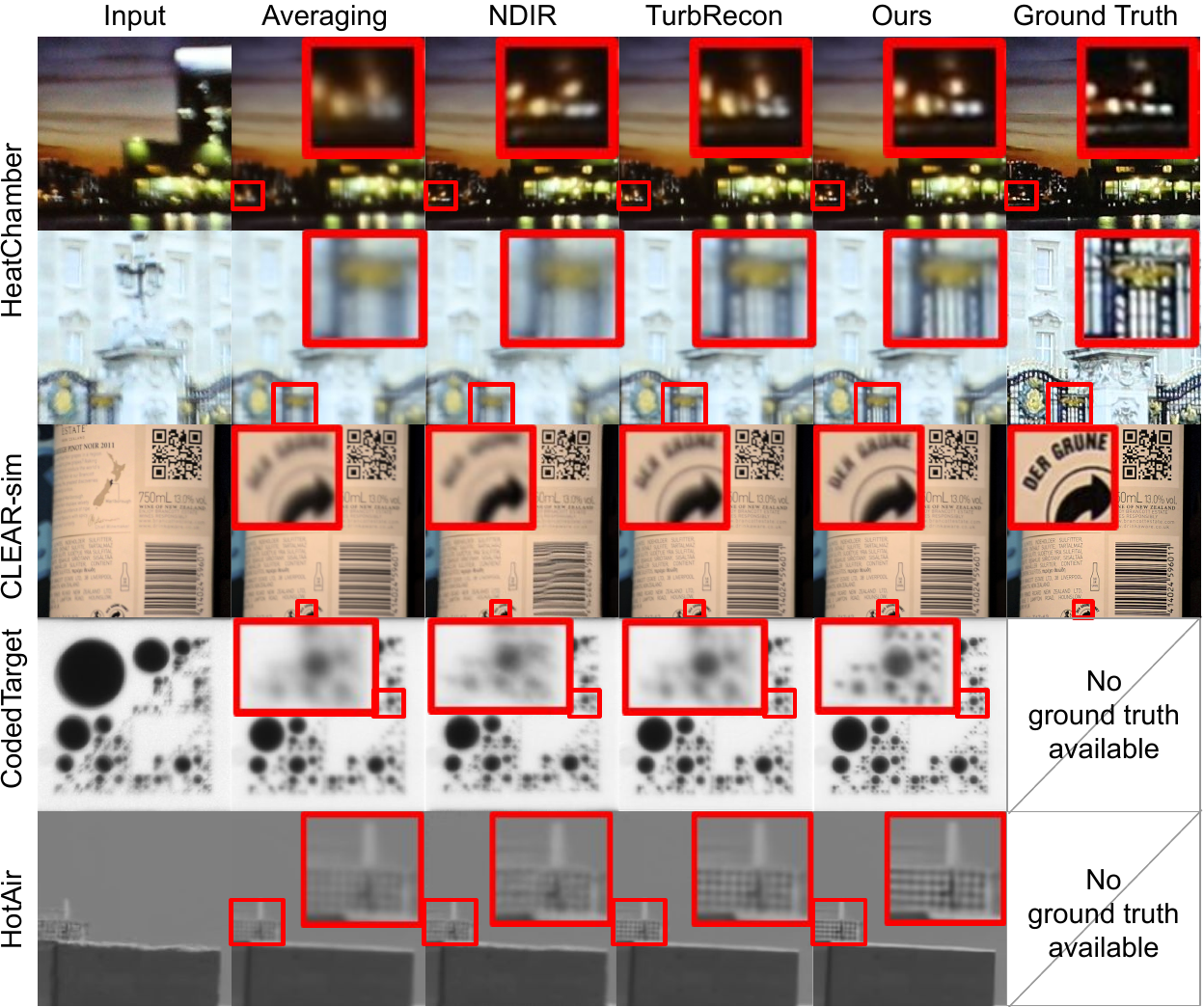}
\caption{\sl\small {\bf Representative qualitative results on various datasets.} Our approach yields crisper and less distorted reconstructions. Note that enhanced registration not only results in reduced distortion, but also mitigates blur caused by misalignment.} 
    \label{fig:results}
    \vspace{-1em}
\end{figure*}

\textbf{Qualitative Results. }
Fig.~\ref{fig:results} presents qualitative results from both datasets. The images are zoomed-in to highlight intricate reconstruction details. Our approach consistently produces reconstructions with minimal distortion compared to the baselines, exhibiting a crisper appearance. It is important to note that we maintain consistent parameters with TurbRecon in optical flow and deconvolution, ensuring that the enhanced reconstruction quality is not a result of parameter selection but stems from superior registration. This also emphasizes that improved registration not only leads to rigid reconstruction with reduced distortion, but also mitigates blur caused by misalignment. A key comparison is in HeatChamber (row 1, Fig.~\ref{fig:results}), where existing methods introduce blurring of lights (highlighted in red) in different directions. Unlike them, we do not rely on an averaging scheme, which enables us to improve on the registration step to yield sharper results. Similar blurring artifacts can be observed in row 2 on the poles of the gate and row 3 (CLEAR-sim) on the logo for both NDIR and TurbRecon.

To provide a comprehensive view, we include two additional examples from the CodedTarget\footnote{https://cvpr2023.ug2challenge.org/} and the HotAir dataset \cite{hirsch2010efficient}. While these sequences lack a ground truth for quantitative evaluation, a visual comparison indicates the consistent superiority of our method.

\begin{table}[t]
\setlength{\tabcolsep}{2.7pt} 
  \centering
  \begin{tabular}{l|c|c|c}
    Method & NDIR~\cite{li2021unsupervised} & TurbRecon~\cite{mao2020image} & Ours-lucky\\ 
    \hline
    PSNR&24.13&24.73&\bf{25.95}\\
    \hline
    SSIM&0.871&0.897&\bf{0.906}\\\hline
    Resolution&$220\times 220$&$1024\times 1024$&$1024\times 1024$
    \end{tabular}
  \caption{\sl\small{\bf Results on CLEAR-sim.} Our method and TurbRecon can handle higher resolutions than NDIR by using optical flow, which avoids the computationally intensive joint optimization. We outperform NDIR and TurbRecon on both PSNR and SSIM.}
     \label{tab:clear}
\end{table}

\section{Discussions and Conclusions}
\label{sec:discussion}
\textbf{Conclusions.} We have presented a simple method; its simplicity makes it easy to extend, and serves as a foundation for more intricate approaches. The crux of it relies on an observation that the mean of the deformation induced by turbulence will converge to the inverse deformation of the flow by the Central Limit Theorem. This assumption makes it possible to register the frames to the latent irradiance to be solved (the ``template'') but without the template. Counter to current trends, our method does not rely on any data-driven processes, but rather an analytical solution. Hence, it is not subject any form of learning bias. Given that the problem is inherently ill-posed, such biases will result in hallucinations, which translates to artifacts and aberrations in the output. Yet, we do make an assumption, which is most generic, making the method widely applicable, and can be seamlessly integrated into mainstream atmospheric turbulence mitigation pipelines. Because of its simplicity, the method is robust and does not exhibit high sensitivity to the choice of frame, i.e. we achieve state-of-the-art perform by using just the 1st frame as the reference frame and the use of simple optical flow. Of course, more informed choices of frame (prior, e.g. sharpest frame) can naturally lead to improved results. We leave the trade-off between performance and generality of the method up to the user.

\textbf{Limitations and future works.} While we are able to achieve state-of-the-art performance on standard benchmarks, admittedly the run time remains a limitation of our approach. The main speed bottleneck of our approach is in the optical flow estimation. The processes of flow inversion and image aggregation involve primarily straightforward interpolations without iterative optimization and contribute negligible computational overhead. Presently, when utilizing the same optimization-based optical flow as TurbRecon, the processing time for 100 frames sized at $220\times 220$ is 17 seconds on an Intel i9-8950hk laptop CPU. When combined with ADMM blind deconvolution, the total processing time is 31 seconds.
We anticipate a substantial speed acceleration by employing contemporary deep learning methods, such as RAFT~\cite{teed2020raft}, as a substitute for Horn \& Schunck. This is expected to reduce the total processing time by half. For reference, TurbRecon requires an additional spatial-time nonlocal averaging module, and the total processing time is 42 seconds. NDIR requires 30 minutes to handle a duration of 10 minutes.

At the core of our method is optical flow, yet mainstream optical flow methods are not specifically tailored for atmospheric turbulence data. Nonetheless, as turbulence simulators advance~\cite{mao2022single}, training a flow customized for air turbulence becomes feasible. This limitation is anticipated to be transformed into a strength, since flow integrates as a plug-and-play module, allowing for straightforward upgrades to our method without alterations to the pipeline.

Finally, we currently do not consider dynamic scenes. Although occlusion induced by motion violates the diffeomorphic assumptions, in Eq.~\eqref{equ:main_deblur} there is in fact no restriction posed on $w$. By strategies like outlier rejection, adjustments can be made to Eq.~\eqref{eq:clt}, enabling the formulation to effectively accommodate scenarios involving either a moving camera or moving objects. Nevertheless, the main goal of this paper is to show that registration can be achieved without the need for an explicit template. Challenges related to dynamic scenes and moving objects will be addressed in future extensions.

{\small
\bibliographystyle{ieee_fullname}
\bibliography{references}
}

\end{document}